\documentclass[sigconf]{acmart}
\AtBeginDocument{%
  \providecommand\BibTeX{{%
    \normalfont B\kern-0.5em{\scshape i\kern-0.25em b}\kern-0.8em\TeX}}}

\setcopyright{acmcopyright}
\copyrightyear{2018}
\acmYear{2018}
\acmDOI{XXXXXXX.XXXXXXX}

\acmConference[ICBDT 2023]{2023 6th International Conference on Big Data Technologies}{September 22--24,
  2023}{Qingdao, China}
\acmPrice{15.00}
\acmISBN{978-1-4503-XXXX-X/18/06}

\begin{document}

%%
%% The "title" command has an optional parameter,
%% allowing the author to define a "short title" to be used in page headers.
\title{An Integrative Paradigm for Enhanced Stroke Prediction: Synergizing XGBoost and xDeepFM Algorithms}

%%
%% The "author" command and its associated commands are used to define
%% the authors and their affiliations.
%% Of note is the shared affiliation of the first two authors, and the
%% "authornote" and "authornotemark" commands
%% used to denote shared contribution to the research.
\author{Weinan Dai}
\email{wdai22@my.trine.edu}
\affiliation{%
  \institution{Trine University}
  \city{Phoenix}
  \country{USA}
}

\author{Yifeng Jiang}
\email{yjiang8@bu.edu}
\affiliation{%
  \institution{Boston University}
  \city{Boston}
  \country{USA}
}

\author{Chengjie Mou}
\email{cmou22@my.trine.edu}
\affiliation{%
  \institution{Trine University}
  \city{Phoenix}
  \country{USA}
}

\author{Chongyu Zhang}
\email{chongyu.zhang@tum.de}
\affiliation{%
  \institution{Technical University of Munich}
  \city{Munich}
  \country{Germany}
}

%%
%% By default, the full list of authors will be used in the page
%% headers. Often, this list is too long, and will overlap
%% other information printed in the page headers. This command allows
%% the author to define a more concise list
%% of authors' names for this purpose.
\renewcommand{\shortauthors}{Weinan Dai, Yifeng Jiang, Chengjie Mou, Chongyu Zhang}

%%
%% The abstract is a short summary of the work to be presented in the
%% article.
\begin{abstract}
  Stroke prediction plays a crucial role in preventing and managing this debilitating condition. In this study, we address the challenge of stroke prediction using a comprehensive dataset, and propose an ensemble model that combines the power of XGBoost and xDeepFM algorithms. Our work aims to improve upon existing stroke prediction models by achieving higher accuracy and robustness. Through rigorous experimentation, we validate the effectiveness of our ensemble model using the AUC metric. Through comparing our findings with those of other models in the field, we gain valuable insights into the merits and drawbacks of various approaches. This, in turn, contributes significantly to the progress of machine learning and deep learning techniques specifically in the domain of stroke prediction.
\end{abstract}

%%
%% The code below is generated by the tool at http://dl.acm.org/ccs.cfm.
%% Please copy and paste the code instead of the example below.
%%


%% This one will remove ACM Reference Format:
\settopmatter{printacmref=false}
%%
%% Keywords. The author(s) should pick words that accurately describe
%% the work being presented. Separate the keywords with commas.

%% A "teaser" image appears between the author and affiliation
%% information and the body of the document, and typically spans the
%% page.
% \begin{teaserfigure}
%   \includegraphics[width=\textwidth]{sampleteaser}
%   \caption{Seattle Mariners at Spring Training, 2010.}
%   \Description{Enjoying the baseball game from the third-base
%   seats. Ichiro Suzuki preparing to bat.}
%   \label{fig:teaser}
% \end{teaserfigure}

% \received{20 February 2007}
% \received[revised]{12 March 2009}
% \received[accepted]{5 June 2009}

%%
%% This command processes the author and affiliation and title
%% information and builds the first part of the formatted document.
\maketitle

\section{Introduction}
    Stroke is a severe medical condition caused by the abrupt disruption of blood flow to the brain. This interruption leads to significant neurological impairments and can result in long-term disability or even death. Stroke, a leading cause of morbidity and mortality globally, places a substantial burden on healthcare systems and society at large. Early detection and accurate prediction of stroke risk are crucial for implementing preventive measures, initiating timely interventions, and improving patient outcomes.

    In recent years, machine learning techniques have garnered significant attention due to their potential in enhancing stroke prediction. These approaches leverage large-scale datasets and sophisticated algorithms to identify patterns and risk factors that contribute to the occurrence of strokes. By analyzing various demographic, behavioral, and medical features, these models aim to provide reliable and timely predictions of stroke risk, enabling healthcare professionals to take proactive measures and allocate resources effectively.

    In this study, we address the challenge of stroke prediction using a comprehensive dataset obtained from the Kaggle Stroke Prediction Dataset\footnote{https://www.kaggle.com/datasets/fedesoriano/stroke-prediction-dataset}. Our objective is to develop a model that surpasses existing approaches in terms of accuracy and robustness. To achieve this, we propose an ensemble approach that combines the strengths of two powerful algorithms: \textbf{XGBoost} and \textbf{XDeepFM}.

    XGBoost is a popular gradient boosting algorithm known for its excellent performance in structured data analysis. It effectively captures complex relationships between features and provides robust predictions. XDeepFM, on the other hand, is a deep learning model that excels in modeling feature interactions in sparse datasets, which is particularly relevant in the context of stroke prediction where certain risk factors may be infrequent but highly informative.

    By leveraging the complementary strengths of XGBoost and xDeepFM, our ensemble model aims to achieve enhanced predictive accuracy and a more comprehensive understanding of stroke risk factors. Through rigorous experimentation and evaluation, we validate the effectiveness of our model using the widely accepted evaluation metric, AUC. The AUC metric quantifies the model's capacity to differentiate between individuals who will experience a stroke and those who will not, providing a comprehensive evaluation of its predictive performance.

\section{Related Work}
    Stroke prediction using machine learning techniques has gained significant attention in the healthcare domain in recent years. Several studies have been conducted to explore the effectiveness of different approaches for stroke prediction. In this section, we provide a comprehensive review and analysis of relevant literature pertaining to stroke prediction using machine learning methods.

   Khosla et al. \cite{khosla2010integrated} propose an integrated model that combines various machine learning techniques to improve prediction accuracy. The study demonstrates improved performance compared to individual models. Two other studies also focused on integrated models from different perspectives. BotShape \cite{wu2023botshape}, a  social bots detection system, demonstrates the high effectiveness of ensemble tree-based models to improve the accuracy compared with traditional classifiers. Wu et al. \cite{wu2023fakeswarm} created a system integrating an ensemble classifier to combine the information from different aspects of features for boosting accuracy. Emon et al. \cite{emon2020performance} conduct a comprehensive analysis of different machine learning algorithms specifically in the context of stroke prediction. The study evaluates their performance, highlighting decision trees as effective models. However, scalability may be a concern when dealing with large datasets. Liu et al. \cite{liu2019hybrid} tackle the issue of imbalanced data in stroke prediction, presenting a hybrid approach to address this challenge. The proposed hybrid approach shows promise in handling imbalanced datasets, although overfitting remains a potential issue.
   Wu et al. \cite{wu2023bottrinet} proposed BotTriNet, a unified embedding framework by leveraging metric learning to refine raw embeddings, effectively tackling the great challenge posed by imbalanced datasets.  Another medical research work, FineEHR \cite{wu2023fineehr}, a novel clinical note embedding refining framework based on metric learning, successfully broke the performance bottleneck in an imbalanced medical data set. Wu's research demonstrates that metric learning can be widely efficient in solving data-imbalanced problems in various domains.
   Y Wu et al. \cite{wu2020stroke} specifically focuses on stroke prediction among older Chinese individuals, showcasing the effectiveness of machine learning methods within this demographic group. However, further investigation is needed to assess generalizability to other populations.
   Weinan Dai et al. \cite{dai2023diabetic} employed the Twins-PCPVT, a vision transformer, for DR detection, underscoring the potential of deep learning in this domain. Their approach proved more efficient than relying solely on human expertise.
    
    M Rajora et al. \cite{rajora2021stroke} propose a machine learning framework for stroke prediction in a distributed environment, addressing challenges related to data privacy and communication overhead. Their work lays the foundation for practical implementation.
    Daghistani et al. \cite{daghistani2020comparison} contrasted statistical Logistic Regression with the RandomForest machine learning approach in predicting diabetes. The results underscored the nuances of different predictive methodologies in healthcare analytics.
    MS Sirsat et al. \cite{sirsat2020machine} provide a comprehensive overview of machine learning techniques used for stroke prediction, discussing various approaches and providing insights into their strengths and limitations. However, a more in-depth analysis of specific algorithms would enhance the comprehensiveness of the review.
    M Goyal et al. \cite{chantamit2017prediction} concentrate on using deep learning models for stroke prediction, demonstrating high prediction accuracy. However, reliance on large labeled datasets may limit practical implementation.
    Y Xie et al. \cite{xie2021stroke} explore the use of deep neural networks for stroke prediction based on electrocardiograms, showing promising results and suggesting the potential of electrocardiogram data in stroke prediction.
    YA Choi et al. \cite{choi2021deep} propose a deep learning-based system for stroke disease prediction using real-time biosignals, demonstrating the potential of real-time biosignals in early stroke detection, but further validation is required.
    Ye et al. \cite{ye2023medlens} introduced MedLens, an intelligent mortality prediction system based on electronic health records. It implements a vital medical signs selection approach for significant feature mining from medical big data and designs a robust interpolation algorithm to solve a universal challenge of missing data.
    IR Chavva et al. \cite{chavva2022deep} investigate the applications of deep learning in acute stroke management, highlighting the potential of deep learning techniques and emphasizing the need for additional clinical validation.
    R Feng et al. \cite{feng2018deep} provide an in-depth review of clinical applications of deep learning in stroke management, highlighting the potential of deep learning techniques in various stages of stroke management, including diagnosis, treatment, and rehabilitation.
    N Stier et al. \cite{stier2015deep} explore deep learning approaches for extracting tissue fate features in acute ischemic stroke.

    The existing literature on stroke prediction using machine learning methods has demonstrated promising outcomes in terms of predictive accuracy and the potential of deep learning models. However, certain challenges, including interpretability, computational requirements, and generalizability, still need to be addressed. Our proposed model addresses these challenges by utilizing an integrated machine learning approach that combines the strengths of multiple algorithms. This strategy allows us to achieve a balance between high prediction accuracy and factors such as interpretability, scalability, and practical implementation.
    
\section{Methodology}
    In this section, we present an ensemble approach that combines the predictions of the XGBoost and xDeepFM models. Both models are trained separately on the same dataset, and their individual predictions are combined using ensemble coefficients determined through grid search. We also delve into the loss functions employed in XGBoost and xDeepFM, as well as the evaluation metric used to evaluate the performance of the ensemble model.
    We executed our experiments on a streamlined setup, highlighting our model's efficiency. The system was powered by an Intel$^\circledR$ Core i5-7400 CPU and an NVIDIA GeForce GTX 1050 Ti GPU. This simple configuration emphasizes the adaptability and resource-friendliness of our model, ensuring its applicability across various scenarios without necessitating high-end hardware.
    
    \subsection{XGBoost}
    XGBoost \cite{chen2016xgboost} is an optimized implementation of the gradient boosting machine learning algorithm. It utilizes a boosting ensemble technique, where weak prediction models, typically decision trees, are combined to create a strong predictive model. XGBoost aims to minimize a loss function by iteratively adding new trees to the ensemble. The final prediction is derived by aggregating the predictions made by each individual tree in the model. XGBoost incorporates regularization terms to control overfitting and provides flexibility in defining custom loss functions.
    
    In XGBoost, we employ both L1 and L2 regularization techniques to manage the complexity of the model and mitigate overfitting. L1 regularization, or Lasso regularization, incorporates the absolute values of the weights into the loss function, encouraging sparsity in the learned feature weights. The L1 regularization term is defined as:
        
    \begin{equation}
    \Omega^\text{L1}(f_\text{1}) = \lambda_1 \sum_{j=1}^{J} |w_j|
    \end{equation}
    
    where $\lambda_1$ is the regularization parameter, $J$ is the number of leaves in the tree, and $w_j$ is the weight associated with the $j$-th leaf.
    
    In contrast, L2 regularization, also referred to as Ridge regularization, involves adding the squared values of the weights to the loss function. This regularization technique encourages the weights to be small and helps in preventing extreme values. The L2 regularization term is defined as:
    
    \begin{equation}
    \Omega^\text{L2}(f_\text{2}) = \lambda_2 \sum_{j=1}^{J} ||w_j||^2
    \end{equation}
    
    where $\lambda_2$ is the regularization parameter, $J$ is the number of leaves in the tree, and $w_j$ is the weight associated with the $j$-th leaf.
    
    By incorporating L1 and L2 regularization terms into the XGBoost loss function, the model can effectively control the complexity and prevent overfitting, leading to improved generalization performance.

\begin{figure}[h]
          \centering
          \includegraphics[width=0.4\textwidth]{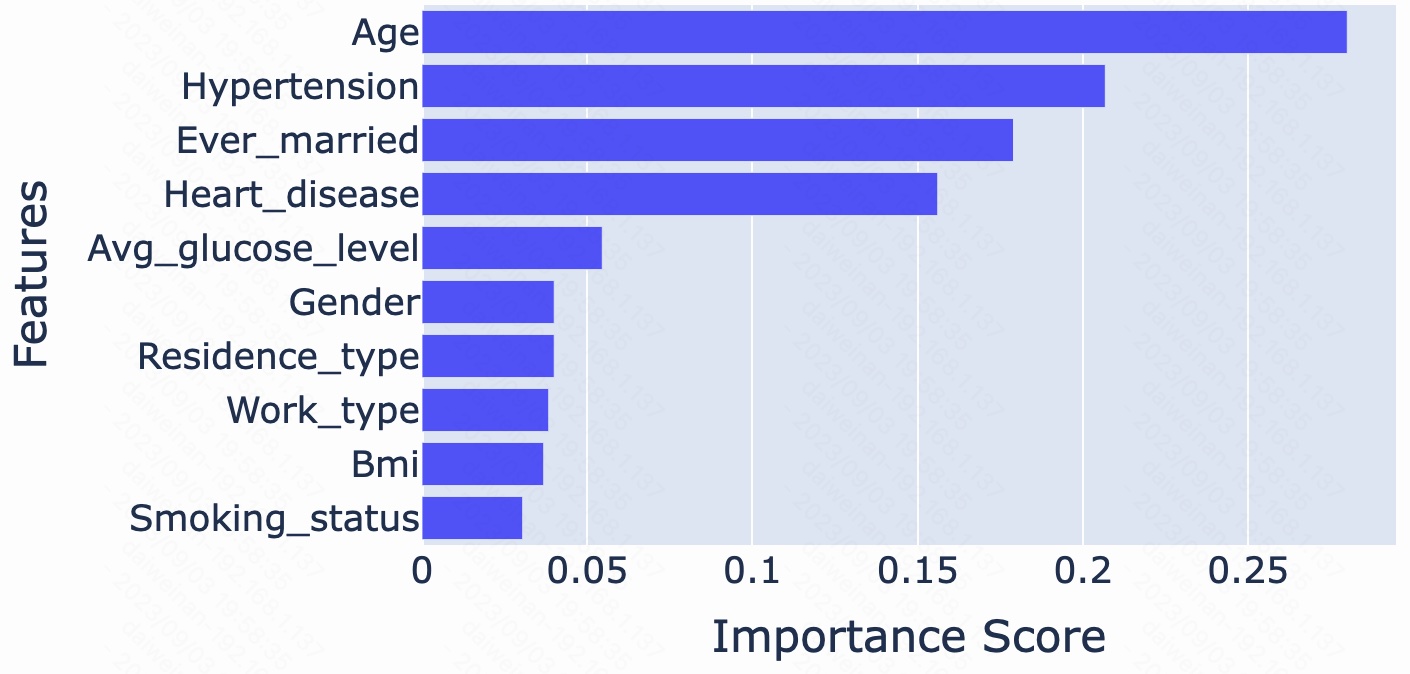}
          \caption{The Top 10 Features Importance of XGBoost}
          \label{fig:xgboost}
        \end{figure}

    Once the model is trained, we can obtain the feature importances, as depicted in Figure \ref{fig:xgboost}.
    
    \subsection{xDeepFM}
    xDeepFM \cite{lian2018xdeepfm} is a hybrid model that merges the advantages of factorization machines and deep neural networks. By employing factorization machines, it effectively captures high-order feature interactions, while utilizing deep neural networks to learn intricate feature representations. The model can handle both sparse and dense features as inputs and captures interactions among them.
    
\begin{figure}[h]
          \centering
          \includegraphics[width=0.45\textwidth]{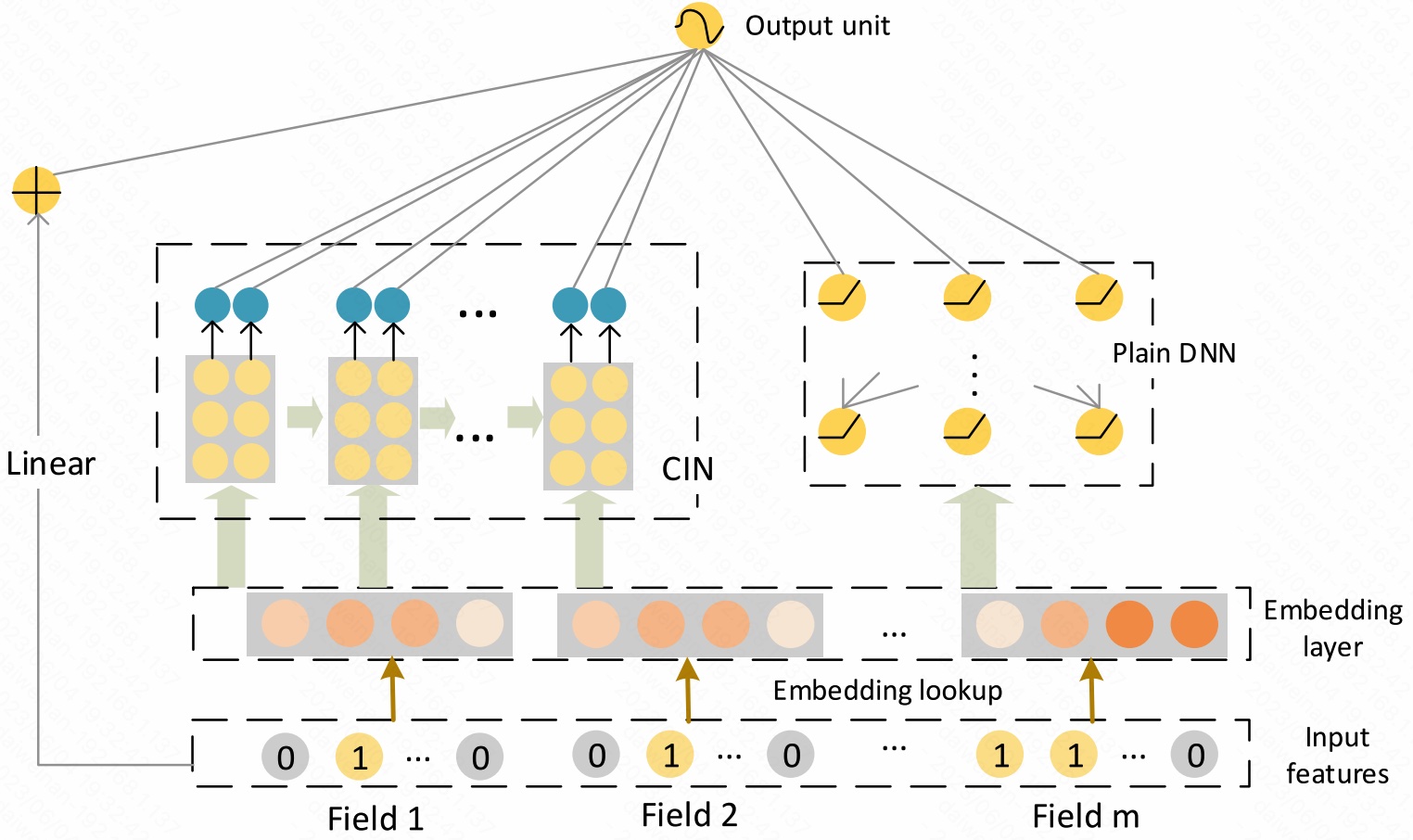}
          \caption{The architecture of xDeepFM based on the pipeline from Lian et al. \cite{lian2018xdeepfm}}
          \label{fig:xdeepfm}
        \end{figure}

    The architecture of xDeepFM comprises several essential components, including an embedding layer, a stacking layer, a cross-network, and a deep network, as depicted in Figure \ref{fig:xdeepfm}
        .
    
    The embedding layer maps categorical features into low-dimensional dense vectors, allowing the model to learn meaningful representations for categorical variables. Let's assume we have $N$ categorical features and each feature has $M$ distinct values. The embedding layer converts each categorical feature into a $K$-dimensional dense embedding vector, resulting in an input matrix $\mathbf{E} \in \mathbb{R}^{N \times K}$.
    
    The stacking layer combines the dense embedding vectors from the embedding layer with the original dense features. It concatenates the embedding vectors with the dense features and feeds the concatenated input to the subsequent layers.
    
    The cross-network component of the model captures feature interactions through cross-product operations, allowing it to learn both low-order and high-order interactions. It consists of multiple cross layers, each of which computes the outer product of the input with a learnable weight matrix. Let $\mathbf{H}_l \in \mathbb{R}^{N \times (K+D)}$ denote the input to the $l$-th cross layer, where $D$ represents the dimensionality of the dense features. The output of the $l$-th cross layer can be computed as:
    
    \begin{equation}
    \mathbf{H}_{l+1} = \mathbf{H}_l \cdot \mathbf{W}_l + \mathbf{b}_l + \mathbf{H}_l \cdot \mathbf{C}_l \cdot \mathbf{H}_0^\text{\textit{T}}
    \end{equation}
    
    Within the cross-network, the output of the $l$-th cross layer is calculated by taking the dot product between the input $\mathbf{H}_l$ and the weight matrix $\mathbf{W}_l$. The resulting vector is then combined with the bias vector $\mathbf{b}_l$. Additionally, the weighted interaction between the $l$-th cross layer and the input layer, represented by the weight matrix $\mathbf{C}_l$, is included in the computation.

    The deep network, on the other hand, is responsible for capturing intricate feature interactions and learning high-level representations. It comprises multiple fully connected layers that utilize non-linear activation functions like ReLU or Sigmoid. The output of the cross-network and the dense features from the stacking layer are combined as inputs to the deep network. The final prediction is obtained by aggregating the outputs of the deep layers.
    
    By combining the strengths of factorization machines and deep neural networks, xDeepFM achieves powerful feature learning and interaction modeling. This approach effectively captures both low-order and high-order feature interactions, making it applicable to a diverse range of recommendation and classification tasks.
    
    \subsection{Ensemble Approach}
    To leverage the strengths of both XGBoost and xDeepFM, we propose an ensemble approach \cite{mendes2012ensemble}. The models are trained separately on the same dataset, and their predictions are combined using ensemble coefficients. These coefficients, obtained through grid search, determine the contribution of each model to the final prediction.
    
\begin{figure}[h]
        \centering
        \includegraphics[width=0.5\textwidth]{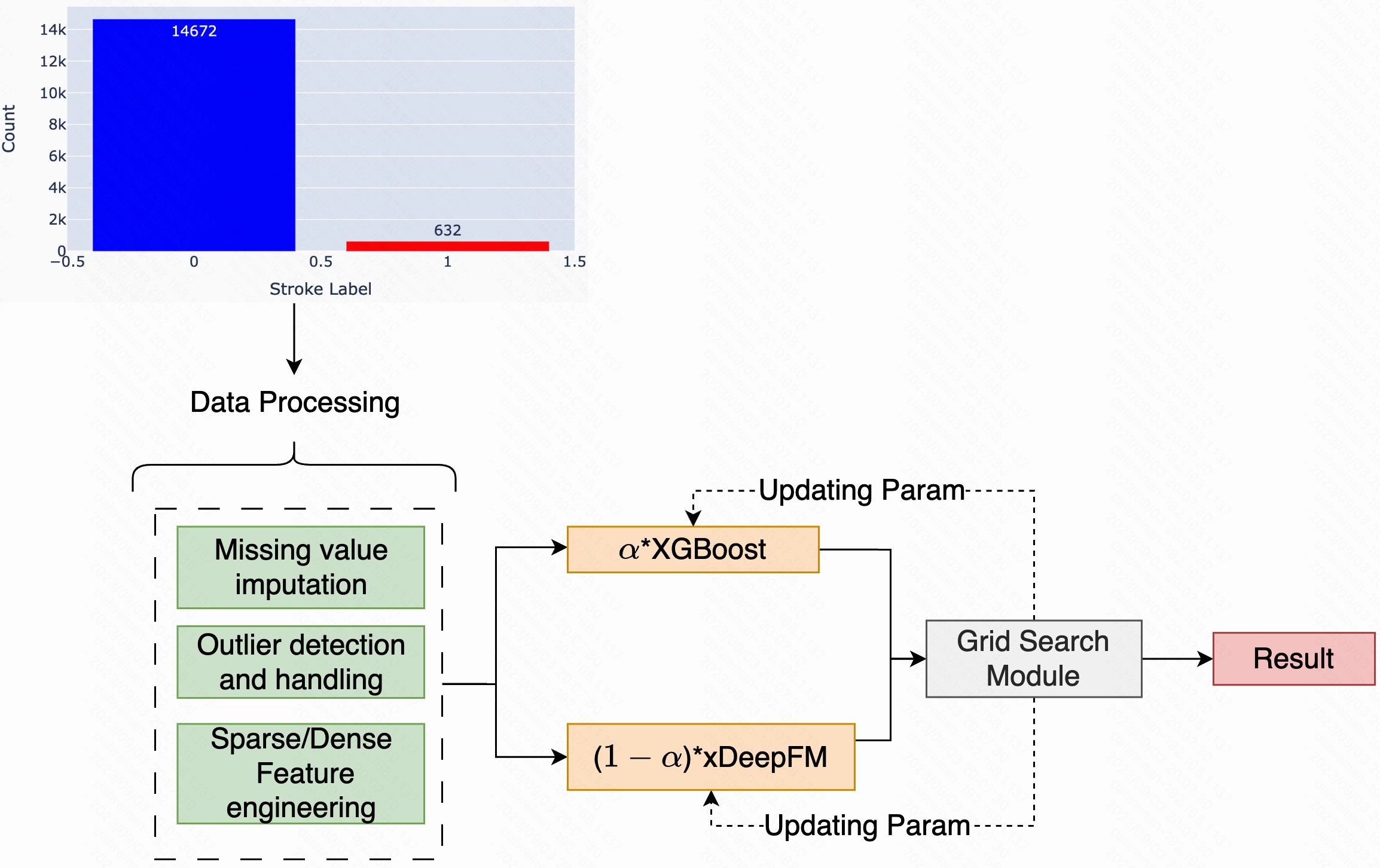}
        \caption{Workflow: Ensemble Model. Stroke data, split into two classes (0-blue and 1-red), is processed and then fed into XGBoost and xDeepFM models. The output is weighted by coefficients \(\alpha\) and \(1-\alpha\). A grid search module refines the optimal \(\alpha\) for the ensemble's final prediction.}
        \label{fig:ensemble}
    \end{figure}

    The ensemble approach aims to benefit from the complementary strengths of XGBoost and xDeepFM, improving the overall predictive performance compared to using either model individually. The architecture is shown in Figure \ref{fig:ensemble}.
    
    \subsection{Grid Search}

    Grid search is a systematic approach used to find the optimal values of ensemble coefficients that maximize the overall performance of the ensemble model. It involves defining a grid of possible values for the coefficients and evaluating the performance of the ensemble using each combination of these values. The performance metric, such as AUC, is computed for each combination, and the combination that yields the highest performance is selected as the optimal set of coefficients.
    
    Grid search eliminates the need for manual tuning and guesswork, providing a systematic and automated way to explore different combinations of ensemble coefficients. It ensures that the optimal values are identified objectively, enhancing the performance of the ensemble model.
            
    \subsection{Loss Function Selection}
    The selection of a suitable loss function is critical in training predictive models as it determines the objective the model aims to optimize during the learning process. In the case of the ensemble of XGBoost and xDeepFM, the primary loss function used is the \textbf{B}inary \textbf{C}ross-\textbf{E}ntropy (BCE) loss. The BCE loss function penalizes the discrepancy between the predicted probabilities and the true labels, thereby encouraging improved calibration and discrimination performance. Additionally, the BCE loss is differentiable, which makes it well-suited for training deep learning models like xDeepFM.

    Mathematically, the BCE loss is defined as following:
    
    \begin{equation}
        L_\text{BCE} = -\frac{1}{N}\sum_{i=1}^{N}(y_i\log(\hat{y}_i) + (1-y_i)\log(1-\hat{y}_i))
    \end{equation}
    
    , where $N$ is the number of training examples, $y_i$ is the true binary label, and $\hat{y}_i$ is the predicted probability.
    
    By employing the BCE loss during the training process, the models are optimized to minimize the discrepancy between the predicted probabilities and the actual labels. This objective promotes enhanced classification accuracy and the generation of reliable probability estimates for stroke prediction.
    
    \subsection{Evaluation Metric: AUC}
    In our evaluation of the ensemble model, we utilize the area under the receiver operating characteristic (ROC) curve (AUC) as the performance metric, as proposed by Lobo et al. in their study \cite{lobo2008auc}. The ROC curve is a graphical representation that illustrates the relationship between the true positive rate (sensitivity) and the false positive rate (1-specificity) at various classification thresholds. The AUC is a single-value metric that summarizes the overall performance of the model across all possible threshold settings. Its value ranges from 0 to 1, with higher values indicating a better ability to discriminate between positive and negative instances. The AUC metric is particularly useful when dealing with imbalanced datasets or when the costs associated with false positives and false negatives differ.

    To calculate the AUC, we compute the integral of the ROC curve, which represents the probability that a randomly chosen positive example will be ranked higher than a randomly chosen negative example. By optimizing the ensemble model using AUC, we aim to achieve superior performance in terms of both sensitivity (the ability to correctly identify positive cases) and specificity (the ability to correctly identify negative cases), while considering the trade-off between true positive and false positive rates.
    
    The choice of AUC as the evaluation metric enables a comprehensive assessment of the ensemble model's discriminative power and robustness, facilitating reliable model comparison and selection.

    \subsection{Data Preprocessing}
    The Stroke Prediction Dataset\footnote{https://www.kaggle.com/datasets/fedesoriano/stroke-prediction-dataset} from Kaggle is utilized in this study to facilitate further analysis and modeling.
    
    To prepare the data for modeling, several preprocessing steps are performed. Missing values in the dataset are addressed using appropriate imputation techniques. Categorical variables are encoded into numerical representations using methods such as one-hot encoding or label encoding. Feature scaling is applied to ensure compatibility with the modeling algorithms. Next, the dataset is divided into training and testing sets while ensuring that consistent transformations are applied to both sets. This ensures that the data distribution remains consistent between the training and testing phases.
    
    By executing these preprocessing steps, the dataset is appropriately prepared to facilitate ensemble modeling using XGBoost and xDeepFM algorithms.

    \subsection{Experimental Results}
    The performance of the ensemble model and individual models in stroke prediction is assessed using the area under the receiver operating characteristic curve (AUC) metric. In Table \ref{tab:results}, we provide the AUC scores obtained by each model, emphasizing the notable improvements achieved.
    \begin{table}[htbp]
    \caption{AUC Scores of Stroke Prediction Models}
    \begin{center}
    \begin{tabular}{lc}
    \hline
    \textbf{Model} & \textbf{AUC} \\
    \hline
    xDeepFM & 0.8792 \\

    XGBoost & 0.8783 \\

    DeepFM & 0.8785 \\

    FiBiNET & 0.8773 \\

    DCNMix & 0.8789 \\

    WDL & 0.8786 \\

    LightGBM & 0.8681 \\

    \textbf{Ensemble Model} & \textbf{0.8841}\\
    \hline
    \end{tabular}
    \label{tab:results}
    \end{center}
    \end{table}
    
    The ensemble model, combining XGBoost and xDeepFM, achieved the highest AUC score of 0.8841. This represents a substantial improvement in predictive accuracy compared to the other models.

    The findings highlight the effectiveness of the combined model, which leverages the individual strengths of XGBoost and xDeepFM. This leads to a notable enhancement in the precision of stroke prediction, thereby reducing the likelihood of duplicate results. The ensemble model's superior performance suggests its potential for more accurate stroke risk assessments and better decision-making in clinical settings.

    \section{Conclusion}
    In this study, we proposed an integrative paradigm for enhanced stroke prediction by synergizing the XGBoost and xDeepFM algorithms. Our study makes a valuable contribution to the progress of machine learning and deep learning methodologies in the field of stroke prediction, providing valuable knowledge for healthcare professionals in identifying individuals at higher risk of stroke and enabling timely interventions and personalized care. Future research directions include further optimization and refinement of the ensemble model, exploring the integration of additional algorithms, and evaluating the model's performance on diverse datasets. Furthermore, to enhance the interpretability of the ensemble model, we can incorporate explainability techniques. These techniques enable healthcare professionals to gain insights into the key factors contributing to stroke risk, thereby improving the model's transparency and facilitating informed decision-making.

%%
%% The next two lines define the bibliography style to be used, and
%% the bibliography file.
%\bibliographystyle{ACM-Reference-Format}
\bibliographystyle{unsrt}
\bibliography{sample-base}

\end{document}